# Learning morphological operators for skin detection

**Alessandra Lumini[1], Loris Nanni[2,*], Alice Codogno[2], Filippo Berno[2]**

[1] DISI, Università di Bologna, Via dell'Università 50, 47521 Cesena, Italy. Email: alessandra.lumini@unibo.it
[2] DEI - University of Padova, Via Gradenigo, 6 - 35131- Padova, Italy
*Corresponding Author: Loris Nanni, Email: loris.nanni@unipd.it

### Abstract

Human skin detection, i.e. the process of discriminating "skin" and "non-skin" pixel in an image or a video, is a very important task for several applications including face detection, video surveillance, body tracking, hand gesture recognition, and many other. Skin detection has been widely studied from the research community resulting in several methods based on hand-crafted rules or deep learning. In this work we propose a novel post-processing approach for skin detectors based on trained morphological operators. The first step, consisting in skin segmentation, is performed according to an existing skin detection approach, then a second step is carried out consisting in the application of a set of morphological operators to refine the resulting mask. Extensive experimental evaluation, performed considering two different detection approaches (one based on deep learning and a handcrafted one), carried on 10 different datasets confirms the quality of the proposed method. To encourage future comparisons the MATLAB source code is freely available in the GitHub repository: https://github.com/LorisNanni.

## 1. Introduction

Human skin detection is the process of discriminating "skin" and "non-skin" regions in an image or a video. The segmentation of skin regions consists in performing a binary classification of pixels and in executing a fine segmentation to define the boundaries of the skin regions. Skin detection has several applications: face detection, video surveillance, body tracking, hand gesture recognition, human computer interaction, biometric authentication and objectionable content filtering among many others [1]. Skin detection is a challenging problem and has been widely studied from the research community. Before the boom of Deep learning, most approaches were based on skin-color separation or texture features. For example, many methods were based on the assumption that skin color can be recognized from background colors according to some clustering rule in a specific color space [2]: then skin detection is performed according to a fixed decision boundary in a color space. A comparison among different color spaces is reported in [3] where basic models (i.e. RGB, normalized RGB) are compared to perceptual models (i.e. HIS, HSV), perceptual uniform models (i.e. CIE-Lab, CIE-Luv) and orthogonal models (i.e. YCbCr, YIQ). Other works studied the problem in the harder assumption of non-constrained environment [3] training general purpose classifiers (i.e. multilayer perceptron [4], random forest [5], Bayesian classifiers [6]). Another class of approaches is based on image segmentation: pixel neighborhood is evaluated in order to segment regions where human skin is present [7][8]. As occurred in many other pattern recognition problems, starting from last few years, convolutional neural networks (CNN) have been applied in skin segmentation with performance increasingly growing. One of the first approaches using deep learning was [9], where patch-based classification is used instead of pixels. In [10] an end-to-end network for human skin detection is designed by the integration of some recurrent neural networks layers into a fully convolutional neural networks. In [11] the authors propose a inception-like architecture, consisting of convolution and ReLU layers only (without pooling and subsampling layers). Finally in [12] the authors present a comprehensive evaluation of different Convolutional Neural Network (CNN) architectures presenting in-domain and cross-domain experiments to determine the best one for skin detection. The experiments in [13] and in [1] confirm the superiority of deep learning over pixel-based or classical region-based approaches even without labelled training samples on the target domain.

The aim of the present work is to propose a novel post-processing approach for skin detection based on trained morphological operators. The detection method is made of two steps: in the first step, a skin detection approach is performed for skin segmentation, in a second step, a set of morphological operators are applied to refine the results. In this work we evaluate two detection approaches: one based on deep learning and an handcrafted one, which according to [1] are the most performing of each category. An extensive experimentation on 10 different datasets using the testing framework proposed in [1] allows a comparison with several state of the art approaches and confirms that the proposed approach is useful to improve detection performance for both of the tested skin detection approaches. Moreover, we used different training protocols in order to show a clear advantage of our learned morphology approach even without training on the target domain.

The organization of this paper is as follows. In section 2 we detail the proposed approach, including a discussion about the fine-tuning of a convolutional neural network for skin detection, a description of the best handcrafted approach used in this work, and a detailed description of the approaches for learning the best morphological operators to be used in the post-processing phase. In section 3 we discuss the experiments, presenting the testing framework used for comparing with other state-of-the-art approaches. Finally, section 4 includes the conclusions and some future research directions.

## 2. Proposed approach

### 2.1. Deep learning for skin detection

In the last few years several deep learned architectures have been proposed for image segmentation task [14]. J. Long et al. [15] have proposed the first deep learned architecture for image segmentation: Fully Convolutional Network (FCN) is a network containing only convolutional layers which adapts existing architecture (i.e. AlexNet, GoogLeNet) into FCN and uses fine-tuning to the segmentation task. Another performing architecture is SegNet [16], which is based on coding and decoding: the coding network is topologically identical to the 13 convolutional layers in the VGG network, the decoding part is designed to map the low resolution encoder feature map to a full input resolution feature map for pixel level classification. More recently a U-shaped network is proposed [17]: a typical encoder-decoder structure where the encoder is aimed at deciding what the object is, and the decoder at delimiting the pixel position. OR-Skip-Net [18] is a recent architecture developed for skin segmentation, whose main idea is to empower the features by transferring the direct edge information from the initial layer to the end of the network.

Since successful training of deep networks requires many annotated training samples usually existing architecture are used for the encoding part, then the network is fine-tuned to the target problem.

In this work we use the same model proposed in [1], a SegNet architecture modeled from VGG19 and fine-tuned using only 2000 labeled images (the first 2000 images of the ECU dataset [19]). Starting from the pre-trained weights of VGG19 on ImageNet, we have removed the last classification layer in the network, and we have added a new weighed classification layer in order to distinguish between "skin" and "non-skin" pixels. We used inverse class frequency (estimated on the training set) weighting in order to deal with imbalanced classes (i.e. the number of non-skin pixel is larger than skin ones). We have used the Stochastic Gradient Descent as optimizer with a learning rate of 0.001 and the Matlab default momentum value (0.9). Since SegNet is designed to work at fixed image size of 224×224 pixels each image is resized using standard MATLAB nearest-neighbor interpolation method before being processed by SegNet, then the resulting mask is resized back to the original input dimension.

## 2.2. Handcrafted method

Most of handcrafted approaches for skin detection are based on the premise that the skin color can be effectively modeled in different color spaces, which allows segmenting the skin regions in color images. [20]. Several approaches are based on *simple rule-based methods* for distinguish skin or non-skin pixel: they are fast and low computational and work well in applications with controlled acquisition conditions, and uniform background (i.e. [20]). More complex approaches includes *adaptive methods* [21], which are based on the tuning of models to the target problem (i.e. lighting variations, skin tone, background). For example in [22] the authors used an explicit skin model in which the optimal color regions are selected from the color spaces, where the skin color is defined as the union of multiple smaller regions. Since such approaches grant performance advantages, even if at the cost of increased computation time, we consider the adaptive method proposed in [21] in our experiments, which is resulted the best handcrafted stand-alone approach in [1]. The approach, named SA3($\tau$), since it depends on a tuning parameter $\tau$, combines a local skin color model created using a probability map and a spatial analysis approach to fix the boundaries of the skin regions. In a first step, a skin probability map is obtained from the input image using a color pixel-wise detector. From this map the high probability pixels are selected as "skin seeds" for the second step, which consists in finding the shortest routes from each seed to every single pixel, in order to propagate the "skinness" and determine the boundaries of the skin regions. Moreover, textural features are used to refine the skin probability map during the second step.

According to [1] SegNet gained the best performance on the proposed benchmark composed by 10 datasets and SA3($\tau$) is the best handcrafted approach. Anyway, the resulting segmentation masks present some artifacts as shown in figure 1.

In order to improve the quality of segmented images we propose a post-processing method based on the use of morphological operators.

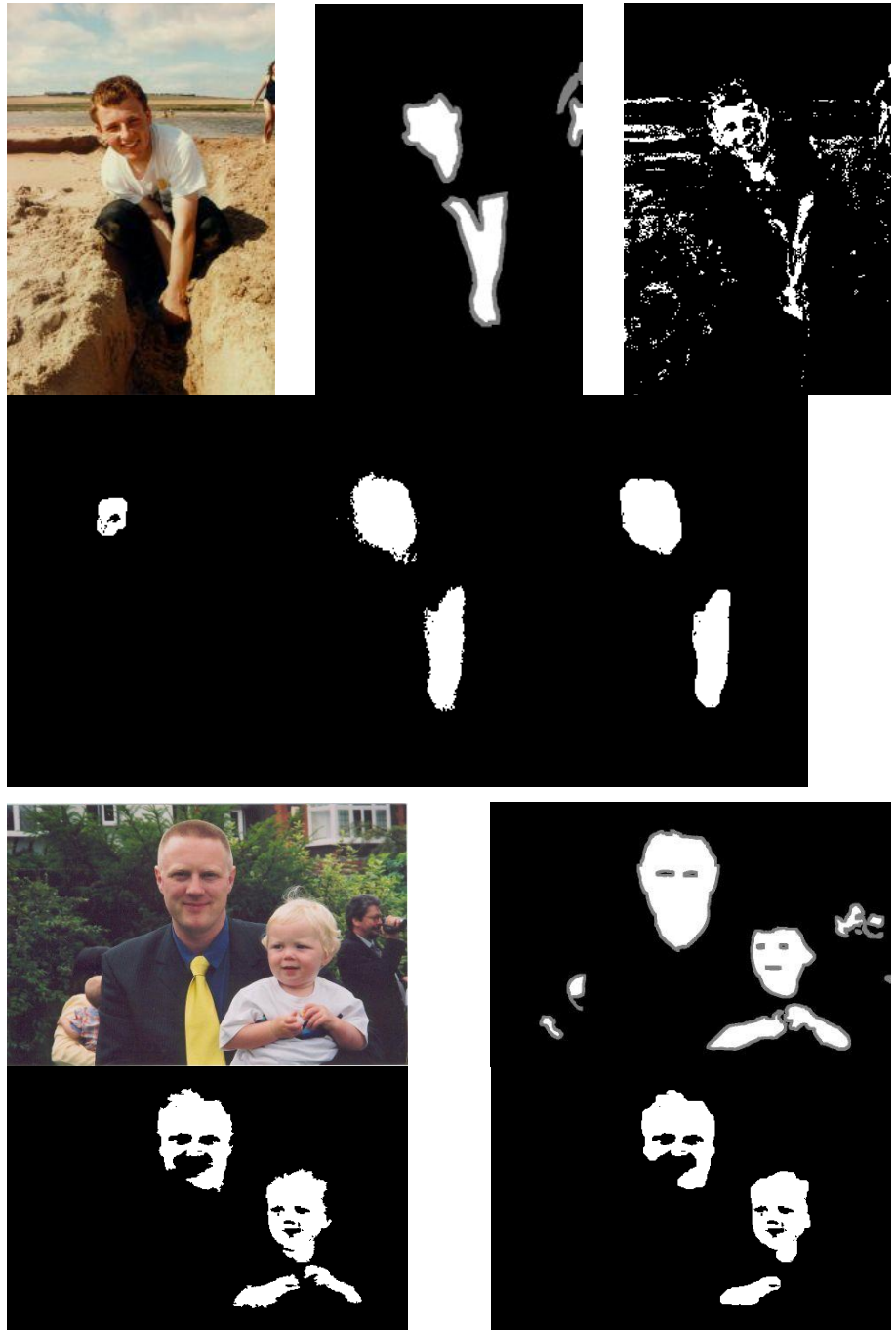

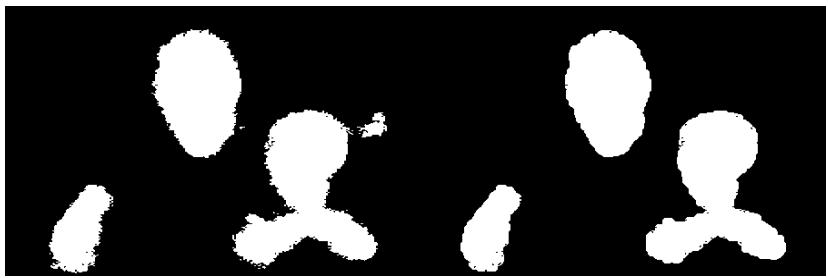

**Figure 1.** Processing of two sample images from the Schmugge dataset. From left to right and from top to bottom: a sample image, its ground truth mask, mask obtained from SA3(τ=50), mask obtained from SA3(τ=50)+Morphological processing, mask obtained from SegNet, mask obtained from SA3(τ=50)+Morphological processing.

### 2.3. Morphological Processing

Analyzing the results regarding the number of true positives (TP) and false positives (FP) obtained from the evaluation of the several skin detection algorithms, we noticed as the most problematic the number of false positives. We have therefore designed a post-processing technique aimed at reducing the quantity of FP, without penalizing excessively the TP.

The proposed method aims to be independent from the skin detection algorithm adopted and it is based on adaptive morphology. Morphological processing [23] can simplify image data while preserving their essential shape characteristics and can eliminate irrelevancies. The operation of dilation and erosion are often used [24], in various combinations, to remove the noise and improve image quality. In the literature there are some recent applications that use morphological processing for skin region detection [25][26], anyway, we noticed that applying the same set of morphological operators to every image in a dataset do not bring the same benefits, and in some cases, the situation has worsened.

To reduce these problems, we tried to discover some features to distinguish among different classes of patterns in order to design an adaptive morphological processing. In our experiments we considered five classes (named A, B, C, D, E in the following), which are discovered according to the following features:

- skin ratio (SR): this is a measure of the percentage of skin in the image and it is calculated as the ratio between the detected skin pixels and total pixels of the image. SR is computed on the original binary mask (named BW) after the use of a closing operator (filling).
- Connected components (CC): it is the number of connected components in the image. CC is computed for images belonging to A, B, C classes after applying an erosion operator on BW, otherwise after using an opening operator.
- Border skin ratio (BSR): it is a measure of skin surface detected in the borders

(1-pixel border). It is used to evaluate the possibility that the background is erroneously considered skin. It is the ratio between the numbers of pixels of skin detected on the top, right, left sides (excluding the bottom) of the image over the total number of pixels of those sides.

In table 1 the rules for classes' definition are reported, according to some threshold values ($a_1, a_2, b_1, b_2, c_1$) which are learned using a training set. Parameter optimization has been performed by grid search on the training set. Each rule is intended as the evaluation of the three conditions in AND. The class E is defined as all the remaining images. Each pattern is sequentially evaluated to each class starting from the first.

The patterns considered to be most "problematic", are those in which the skin detection algorithm assessed the background of the image as skin, from which many FP are produced (class A).

Each pattern is post-processed using a standard sequence of operators, independently from the class they belong to: erosion, opening, dilation and multiply operators. Then for classes A and E an additional closing operator is used. Only for patterns labeled as A, each connected component is singularly evaluated in order to find the one which can be considered as background and thus removed.

The holistic rule to label and remove the background is the following: select the largest component for the pattern obtained from the original BW mask after using appropriate erosion and dilation operators. We are aware that this thumb rule could be improved, anyway it performs well at a low computational cost. In table 2 the morphological operators used in this work are explained, and in table 3 the list of operators used for each class is reported.

**Table 1.** Rules adopted to label a pattern with a certain class

| Class | Semantic Description | Features | | |
|---|---|---|---|---|
| | | SR | CC | BSR |
| A | Background classified as skin | $\geq a_1$ | $< b_1$ | $\geq c_1$ |
| B | Face in the foreground | $\geq a_1$ | $< b_1$ | $< c_1$ |
| C | Single person o little group. | $\geq a_1$ | $\geq b_1$ | - |
| D | Group of people | $> a_2$ & $< a_1$ | $> b_2$ | - |
| E | Unknown | otherwise | otherwise | - |

**Table 2.** Morphological operations performed (We use as structural element a disk with radius 6)

| MATLAB Operators | MATLAB Operators |
|---|---|
| imerode(I,SE) | Erosion: erode mask I with structuring element SE |
| imdilate(I,SE) | Dilation: dilate mask I with structuring element SE |
| imfill (I,'holes') | Fills holes of the input mask I |
| immultiply(I,J) | Multiply two masks pixel-by-pixel |
| bwareaopen (I,P) | Removes small objects (≤P pixels) from I |

**Table 3.** Sequence of morphological operations (MATLAB commands) performed to each class

| Class | MATLAB Operators |
|---|---|
| A | imerode, imdilate, immultiply, imfill, imerode, bwareaopen, imdilate, immultiply |
| B,C,D | bwareaopen, imerode, bwareaopen, imdilate, immultiply |
| E | imfill, imerode, bwareaopen, imdilate, immultiply |

As baseline performance we have checked several sets of morphological operators, we have obtained the best results varying the method proposed in [25], where a set of morphological operators is introduced for face detection. In this work, we have modified them for a generic image[1], the following operations are applied:

1. Morphological closing;
2. Morphological erosion, structuring element (disk) of 10 pixels;
3. Morphological dilation, structuring element (disk) of 8 pixels;
4. Dilated binary image is multiplied with binary image from the segmentation process to maintain the holes.

The pseudo code of the proposed approach is reported in figure 2.

---

[1] http://www.cvip.louisville.edu/old/wwwcvip/education/ECE523/Spring%202011/Lec5.pdf

```
fun PostProcessingOM(BW, a1,a2,b1,b2,c1) {
// Input values: BW is the result of the skin detection
//   a1,a2,b1,b2,c1 are the trained thresholds
  class <- 'E'    // BW is assigned to the default class
  doFill <- 1     // Flag for the opening operator set to 1
  Compute SR on (filled BW) // Skin Ratio
  if (SR >= a1) { // BW belongs to A,B or C
     eBW <- Heavily erode BW to divide connected components in smaller ones
     Compute CC on eBW // Number of connected components
     if (CC < b1) { // BW belongs to A or B
        Compute BSR on eBW // Border skin ratio
        if (BSR >= c1)   class <- 'A' // BW is assigned to A
        else    class <- 'B' // BW is assigned to B
     }
     else class <- 'C' // BW is assigned to C
  }
  else if (a2 < SR < a1) { // BW belongs to D or E
     Compute of CC on copy of BW where cc smaller than 10px are removed
     if (CC > b2)   class <- 'D' // BW is assigned to D
  }
  switch (class) {
      case 'A':
           bgArea <- Determined background area of eBW
           NBW <- Dilated bgArea is removed from BW
      case 'B','C','D': // Same processing for B, C and D
           doFill <- 0
           NBW <- BW
      case 'E':
           NBW <- BW
  }
  if (doFill)
       NBW <- NBW without holes // opening operator is applied
  else
       NBW <- NBW - Holes smaller than 3px
Final processing on NBW: erosion, remove small cc, dilation
...
  NBW <- BW*NBW //immultiply
  return NBW
```

Figure 2. Pseudo-code of the proposed method.

## 3. Experimental Results

The experimental evaluation of the proposed approaches has been performed according to the testing framework proposed by [1], which includes the following 11 datasets (10 used only for testing and one for training):

**ECU** [19] includes 4000 color images. The first half of this datasets is used for training purposes.

**Compaq** [27] is one of the first and most used large skin dataset, in our version we included the 4675 skin images supplied of ground truth.

**UChile** [28] is a small dataset including 103 images acquired in different lighting conditions and with complex background.

**Schmugge** [29] is a collection of 845 images taken from different face datasets.

**Feeval** [30] is a dataset of 8991 frames extracted from 25 online videos of low quality. Here, the performance is calculated averaging the performance of each video by the number of frames (therefore considering each video as a single image in the performance evaluation).

**MCG** [29] contains 1000 images.

**VMD** [31] includes 285 images from several public datasets for human activity recognition.

**SFA** [32] includes face images from different databases.

**Pratheepan** [33] is a small dataset which includes 78 images.

**HGR** [21] is a dataset for gesture recognition. Since the image size is very large, in our experiments the size of the images of HGR2A and HGR2B has been reduced of a factor 0.3.

**FvNF** [34] (Face vs. NonFace) is not a real skin dataset, it is composed by 800 face and 770 non-face images, extracted from Caltech dataset [35]. This dataset is used to evaluate the capability of a skin detector method to detect the presence of a face, therefore the average precision (*AP*) is used as performance indicator.

According to most of the methods proposed in the literature we use the following performance indicators:

- $F_1$ *measure* is the harmonic mean of precision and recall and it is calculated according to the following formula:
  $F_1 = 2tp/(2tp + fn + fp)$
  where *tn* are true negatives, *fn* are false negatives, *tp* are true positives and *fp* are false positives. For the skin detection problem $F_1$ is averaged at pixel level not at image level; in such a way the final indicator is not dependent on the image size in the different databases.
- *AP (Average precision)*, $AP \in [0,100]$ is the area under the precision-recall

curve.

In our experiments we fix the acceptance threshold of SA3($\tau$) to $\tau = 50$ (which is the best performing value reported in [1]) and we maintain the same value of threshold for all the datasets. A fair comparison among different skin approaches is difficult due to the difference in applications and datasets for skin detection. However in [1] a unified framework for standard in evaluation is proposed, made of a unified testing protocol and 10 datasets having different targets and characteristics: acquisition method, target application, illumination conditions. In this work we have validated the proposed algorithm for morphological processing using the 10 datasets described above. Table 4 reports the performance of the following approaches: the two baseline skin detection approaches, i.e. SegNet and SA3(50), the result of the application of a fixed set of morphological operators (BM) to them both, the application of a trained set of morphological operators as described in section 2.3. Parameters' training for morphological operators has been performed according to two different training protocols: leave-one-out dataset (TM) and ECU (EM); in the first case a more accurate parameter selection is performed considering all datasets excluding the testing one, in the other case the first 2000 images of ECU are used for training (the same subset used for deep learning). The last row of Table 4 reports the global rank, calculated as the rank of the average rank on each dataset.

The optimal EM parameters (a1,a2,b1,b2,c1) learned for SA3 and SegNet using all the datasets are:

(0.3, 0.06, 16, 48, 0.55) and (0.3, 0.06, 10, 40, 0.25) respectively.

**Table 4.** Performance of the proposed approach in the 10 dataset using F-measure for the first nine dataset and average precision (AP) in the last.

| Dataset | Method | | | | | | | |
|---|---|---|---|---|---|---|---|---|
| | SA3(50) | +BM | +TM | +EM | SegNet | +BM | +TM | +EM |
| *Feeval* | 0.539 | 0.54 | 0.532 | 0.535 | 0.711 | 0.717 | 0.715 | 0.715 |
| *Prath.* | 0.709 | 0.711 | 0.729 | 0.720 | 0.73 | 0.743 | 0.747 | 0.746 |
| *MCG* | 0.762 | 0.761 | 0.771 | 0.765 | 0.813 | 0.819 | 0.821 | 0.819 |
| *UChile* | 0.625 | 0.626 | 0.627 | 0.625 | 0.802 | 0.808 | 0.809 | 0.809 |
| *Compaq* | 0.647 | 0.654 | 0.652 | 0.653 | 0.737 | 0.745 | 0.75 | 0.744 |
| *SFA* | 0.863 | 0.849 | 0.849 | 0.849 | 0.889 | 0.895 | 0.898 | 0.898 |
| *HGR* | 0.877 | 0.903 | 0.905 | 0.903 | 0.869 | 0.866 | 0.867 | 0.867 |
| *Schmugge* | 0.586 | 0.618 | 0.62 | 0.62 | 0.708 | 0.72 | 0.721 | 0.721 |
| *VMD* | 0.147 | 0.14 | 0.14 | 0.14 | 0.328 | 0.346 | 0.352 | 0.352 |
| *FnF(AP)* | 89.8 | 90.28 | 91.25 | 90.50 | 99.98 | 99.97 | 99.98 | 99.97 |
| Rank | 8 | 7 | 5 | 6 | 4 | 3 | 1 | 2 |

The execution time of our post-processing step is negligible: the elapsed time using BM is 0.015595 seconds for a 224×224 image (i7-7700HQ 2.8 Ghz), while the elapsed time using the learned morphology (TM or EM) is 0.033092 seconds.

To statistically validate our experiments we also perform the Wilcoxon signed rank test [36], a nonparametric test that compares the performance of classifiers by considering the number of wins and losses and the difference in performance on different datasets. The Wilcoxon signed rank test confirms that:

- SegNet+BM outperforms SegNet with p-value 0.0143
- SegNet+TM outperforms SegNet with p-value 0.0058
- SegNet+TM outperforms SegNet-BM with p-value 0.0172

From a comparison among stand-alone approaches in table 4 and from the above results using Wilcoxon signed rank test, the following conclusions can be drawn:

• The performance of almost all the approaches is strongly affected by the quality of the images, therefore varies widely from one dataset to another. As expected, all approaches based on deep learning outperform the hand-crafted ones, anyway each method could improve performing a fine-tuning per dataset.

• As handcrafted approaches are concerned, the selection of an appropriate threshold is crucial for performance but depends of the specific application.

• As deep learned methods are concerned a fine tuning on a specific dataset is crucial: for example, HGR is a dataset for gesture recognition which could be better segmented by a CNN appositely trained for hands/arm segmentation. This is beyond the scope of this work where we maintained the same configuration for a fair comparison on very different applications.

• The use of morphological operators improves the performance of base approaches in almost all the datasets. Moreover, the trained versions of morphology proposed in this work (TM and EM) perform better than the base version (BM).

• A simple training on a small set of images is enough to tune our approach as proved by the very similar performance obtained by EM vs. TM.

A visual inspection of resulting images is useful to analyze the performance of the proposed approach. In figure 3 the resulting of morphological post-processing to 3 different images is shown (the masks where obtained by SegNet approach to images from MCG dataset). In the first case the segmentation is improved by morphological processing, while in the second and third cases the amount of correctly detected skin pixels decreases. A cause is certainly the quality of the original image, which is very low for the second sample, but for the third sample the main reason is that the proposed morphological approach can work only to reduce the number of false positive pixels detected as skin, but cannot handle the false negatives. To handle this problem the proposed approach should work starting from "probability" masks instead of binary masks and using different thresholds to simulate different degree of skinness. This idea can be considered for a future work.

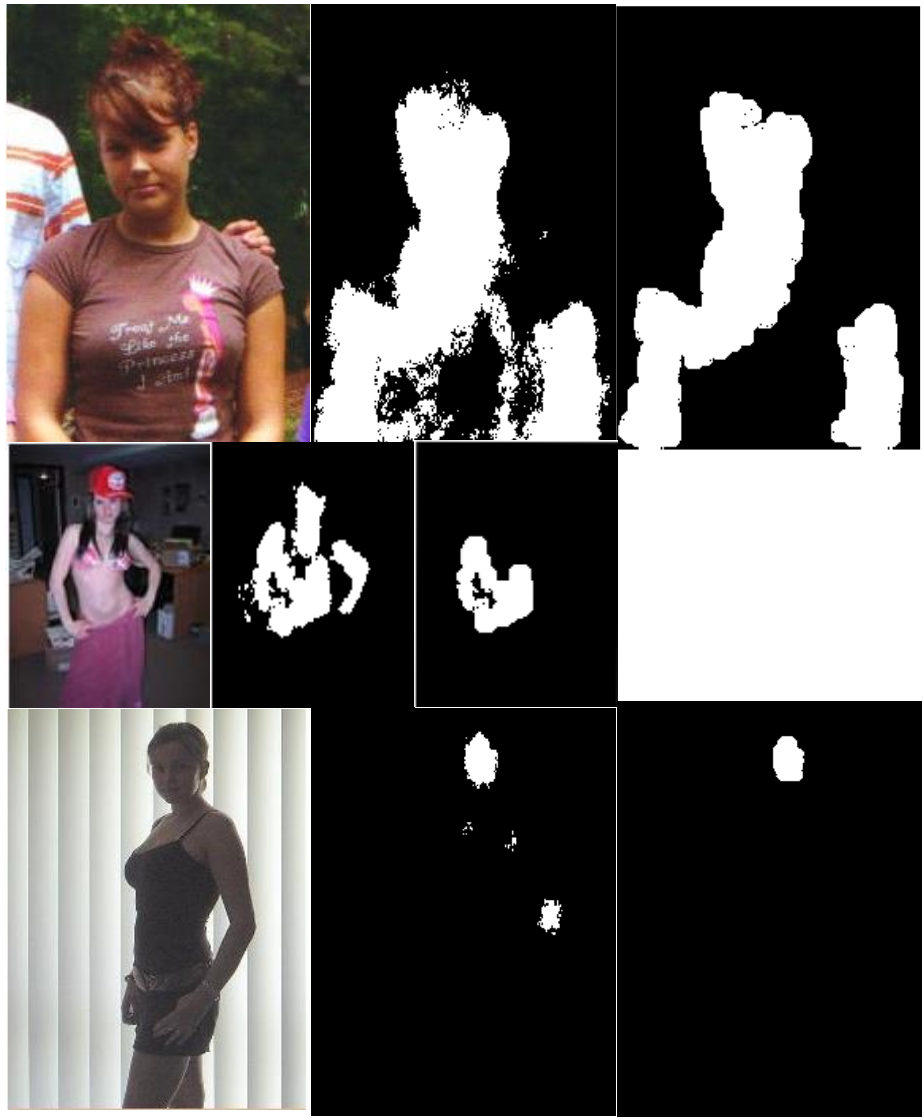

**Figure 3.** Processing of three sample images from MCG dataset. From left to right: RGB image, skin mask by SegNet, skin mask by SegNet+BM.

First Author et al.

## 4. Conclusions

In this work we proposed a novel post-processing approach for improving performance of skin detectors. The post-processing is based on trained morphological operators which are applied on the skin mask obtained by a skin detection approach. The set of morphological operators applied to refine the results are not a priori fixed but selected according to some trained rules. In the present work, even if 5 different classes of images are defined, 3 of these classes are post-processed using the same set of morphological operators. In future work, investigating the possibility to differentiate such processing might be important. The trained morphological approach is evaluated with two well performing skin detection approaches: one based on deep learning and a handcrafted one. An extensive experimentation on 10 different datasets show that the proposed approach is useful to improve detection performance for both tested skin detection approaches. In conclusion, we show that even if deep learning performs very well for the skin detection problem, the use of a hand-crafted post-processing method can further improve the performance.

Future research could examine the processing of low-quality images. Our experiments proved that the performance of SegNet decades for low quality images, and post processing does not help in such cases. A possible reason could be that the network has been trained only on high quality images; in such a case retraining the CNN using data augmentation for simulating low quality images could improve the segmentation performance. Future studies should also aim to replicate results in datasets including images from different races, not considered in this study, as the dataset proposed in [18] (not available at the time of this study)

## Acknowledgements

We would like to acknowledge the support that NVIDIA provided us through the GPU Grant Program. We used a donated TitanX GPU to train CNNs used in this work.

## Conflicts of Interest

We have no conflict of interest to declare